# Development of an accessible 10-year Digital CArdioVAscular (DiCAVA) risk assessment: a UK Biobank study

Nikola Dolezalova [1], Angus B. Reed [1], Alex Despotovic [1,2], Bernard Dillon Obika [1,3], Davide Morelli [1,4], Mert Aral [1], David Plans* [1,5,6]

1. Huma Therapeutics Limited, London, United Kingdom
2. Faculty of Medicine, University of Belgrade, Belgrade, Serbia
3. Barking, Havering and Redbridge University Hospitals NHS Trust, London, United Kingdom
4. Department of Engineering Science, Institute of Biomedical Engineering, University of Oxford, Oxford, United Kingdom
5. Department of Experimental Psychology, University of Oxford, Oxford, United Kingdom
6. Department of Science, Innovation, Technology and Entrepreneurship, University of Exeter, Exeter, United Kingdom

   **\* Corresponding Author:**
   *Email Address:* david.plans@huma.com
   *Mailing Address:* Huma Therapeutics Ltd., Millbank Tower, 21-24 Millbank, London, SW1P 4QP, United Kingdom

**Running Head:** Development of DiCAVA: a UK Biobank study

# Abstract

**Background:** Cardiovascular diseases (CVDs) are among the leading causes of death worldwide. Predictive scores providing personalised risk of developing CVD are increasingly used in clinical practice. Most scores, however, utilise a homogenous set of features and require the presence of a physician.

**Objective:** The aim was to develop a new risk model (DiCAVA) using statistical and machine learning techniques that could be applied in a remote setting. A secondary goal was to identify new patient-centric variables that could be incorporated into CVD risk assessments.

**Methods:** Across 466,052 participants, Cox proportional hazards (CPH) and DeepSurv models were trained using 608 variables derived from the UK Biobank to investigate the 10-year risk of developing a CVD. Data-driven feature selection reduced the number of features to 47, after which reduced models were trained. Both models were compared to the Framingham score.

**Results:** The reduced CPH model achieved a c-index of 0.7443, whereas DeepSurv achieved a c-index of 0.7446. Both CPH and DeepSurv were superior in determining the CVD risk compared to Framingham score. Minimal difference was observed when cholesterol and blood pressure were excluded from the models (CPH: 0.741, DeepSurv: 0.739). The models show very good calibration and discrimination on the test data.

**Conclusion:** We developed a cardiovascular risk model that has very good predictive capacity and encompasses new variables. The score could be incorporated into clinical practice and utilised in a remote setting, without the need of including cholesterol. Future studies will focus on external validation across heterogeneous samples.

## Graphical abstract

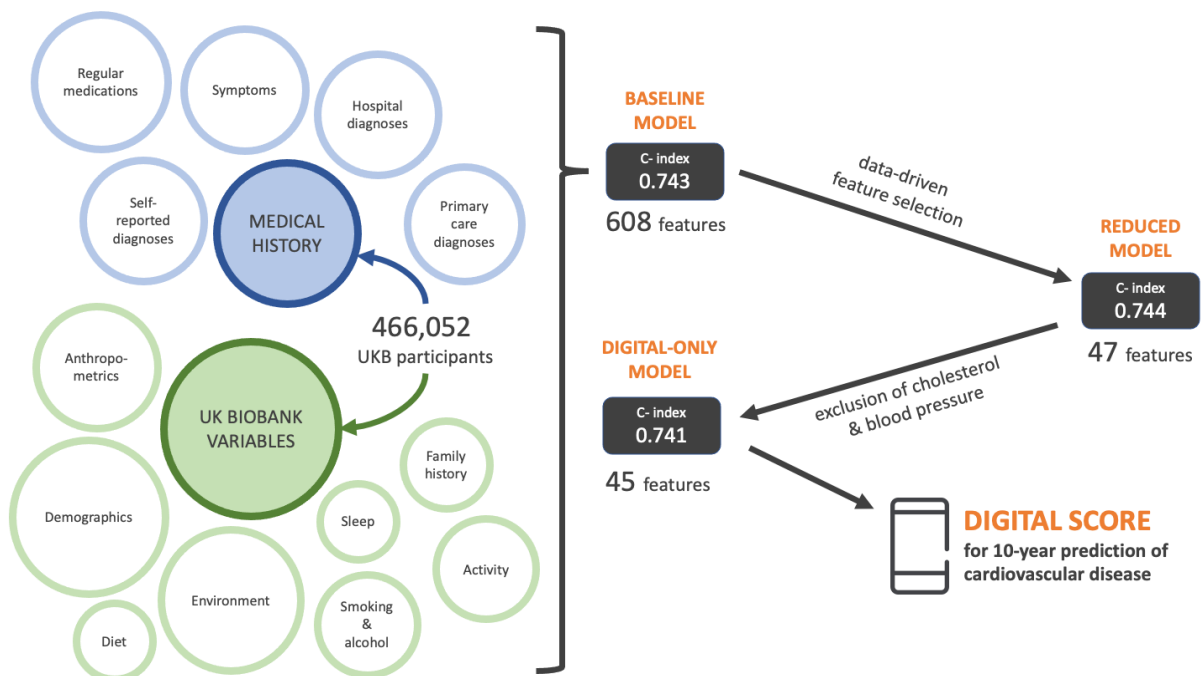

# Introduction

Cardiovascular diseases (CVDs) are the most common cause of mortality globally, with 18.6 million attributed deaths in 2017 (1). The increase of CVD-related mortality in low- and middle-income countries (2), together with increased life expectancy and a growing CVD morbidity worldwide (3), poses a significant challenge in managing this group of diseases. Many risk factors for acquiring CVDs are well-established: hypertension, obesity, diabetes, poor physical activity, hypercholesterolaemia, smoking, alcohol intake, among others (4). Given that up to 70% of cases and deaths from CVDs are attributed to modifiable factors (4), primary prevention of CVD, rather than treatment, has become a mainstay of public health strategies and is extensively described in leading guidelines (5). These strategies can be highly effective in reducing the number of occurrences and, thereby, the corresponding mortality.

The use of risk scores for CVDs in clinical practice is commonplace (6–10). Their primary use is to identify individuals who are at high risk of development of either a fatal and/or non-fatal CVD event in the next 10 years. Their goal is to highlight that risk, so it can be mitigated through either lifestyle adjustment or pharmacological treatment of associated conditions, such as hypertension and hypercholesterolaemia. These interventions have indeed been shown to be cost-effective (11), but the pursuit of better screening programmes and economic assessments is necessary (12).

There are several limitations of the cardiovascular risk scores used in practice. In addition to overestimating the proportion of people in the high-risk category (13), recent systematic reviews show low-quality evidence that use of existing CVD risk scores may have minimal effect on incidence of CVD events (14). Moreover, the majority of existing risk scores utilise a small and homogenous set of nonmalleable factors in their calculation (***Supplementary Table 1***).

Risk factors for CVDs can be easily captured through web-based platforms, with standard practice often integrating a risk model as a function within electronic healthcare records (EHRs) (15). However, more granular insight into cardiovascular health can be achieved through smartphone-based apps and integrated questionnaires, or wearable devices that allow passive and continuous assessments (16). Furthermore, the analysis of such data using modern methods, such as machine learning (ML), allows for deeper insight into the risk factors that contribute to CVDs, capturing non-linear variable interactions uncapturable by classical statistical methods (17).

The primary aim of our study, therefore, is to develop a CVD risk model using both traditional statistical and ML approaches, guided by clinical intuition, which is viable in a remote setting. The secondary goal is the identification of new variables which should be considered by the wider community for incorporation into future CVD risk models to improve their utility. To evidence the value of non-traditional variables, we compare our model to the Framingham risk score that is currently used in clinical practice.

## Methods

**Study design and exclusion criteria**

The UK Biobank (UKB) dataset is a prospective cohort study of 502,488 UK participants (18) recruited between 2006 and 2010. It comprises baseline information from an initial assessment on participants who were subsequently followed up via additional assessment sessions, medical records, and other health-related records. Use of data for this study was approved by UKB (application number 55668).

The sole exclusion criterion was a pre-existing diagnosis of a CVD. Censoring was done at the date of first CVD diagnosis, when lost to follow-up due to death or other reasons, or at the date of last available update on the data (30th September 2020), whichever came first.

**Outcome definition**

Definition of CVD in this study includes myocardial infarction, heart failure, angina pectoris, stroke, and transient ischaemic attack. A diagnosis of CVD was confirmed using several fields of UKB-defined First Occurrences (includes primary care, hospital admissions, death register data, and self-report), Algorithmically-defined Outcomes (hospital admissions, death register data, and self-report), and remaining ICD-10 codes from inpatient records. The full list of UKB fields used for outcome definition can be found in ***Supplementary Table 2***. While there was >75% overlap between the Algorithmically-defined Outcomes and the corresponding fields for First Occurrences, taking into account both fields ensured a more comprehensive selection of CVD cases. Earliest date for any cardiovascular disease on record was used for each participant.

**Variable selection**

For the baseline set of variables, we used UKB variables available for the majority of participants. This dataset was further enriched by including First Occurrences fields for all available ICD-10 (10th revision of the International Statistical Classification of Diseases and Related Health Problems) codes and hospital inpatient records for remaining ICD-10 codes (***Supplementary Table 3***). Diagnoses with less than 0.2% occurrence in the dataset were excluded from the feature list. Considering our use-case, a key determinant was the possibility of its input or assessment using only a smartphone and the possibility to apply these findings to other countries. This led to exclusion of blood tests and other biological measurements which cannot be acquired via smartphone, as well as UK-specific variables such as Townsend index, income, and certain qualification levels. The exceptions were total cholesterol, HDL cholesterol, and blood pressure measurements owing to their significant predictive value in CVD risk.

The majority of predictor variables were used as provided by the UKB, with the addition of several variables which were derived from the data (waist-to-hip circumference ratio, total cholesterol/HDL cholesterol ratio, total alcohol intake). Imputation of missing values was performed by substitution with the mean. Binary variables for pre-existing diseases were derived from a combination of corresponding fields in First Occurrences, Self-reported medical conditions, medications, and in-patient hospital diagnosis, with the exclusion of cases where diagnosis occurred after the assessment date. Categorical and ordinal variables were one-hot encoded and continuous variables were scaled.

**Feature selection and prediction models**

Prior to training the model, the dataset was split into two parts: train dataset (75%) and test dataset (25%), stratified on the outcome. The train dataset was further split into train (75%) and validation (25%) for the purposes of DeepSurv hyperparameter search. There were no statistical differences between the different datasets (data not shown).

Our modelling approach involved the use of Cox Proportional Hazard (CPH) model, implemented in the Python lifelines library (19), and Cox proportional hazards deep neural network (DeepSurv), implemented in the 'pycox' package (20) using PyTorch (21). Using the full set, and later the reduced set of features, optimal hyperparameters were searched among those described in *Table 1* via Tree-Structured Parzen Estimator algorithm (22) from the Optuna Library (23).

| Hyper-Parameter | Search Space |
| --- | --- |
| Activation | LeakyReLU (24), ReLU (25), and SELU (26) |
| Hidden Layers topology | 8, 32, 256, 32x32, 64x64, 128x128, 64x16, 256x32, 32x32x32, 64x64x64 |
| Drop-Out* (27) | [0, 0.9] |
| Weight-Decay* (28) | [0, 20] |
| Batch Normalization (29) | Yes/No |
| Optimizer | Stochastic Gradient Descent, Adam (30) |
| Momentum* (31) | [0,1] |
| Learning Rate | Log distribution on [1e-5, 1] |

*Uniform distributions*

*Table 1: DeepSurv hyper-parameter search space.*

Feature selection was performed using the CPH model to decrease the risk of overfitting and ensure suitability for use in a digital solution. In the first step, univariate CPH models were trained for each of the features in the baseline model and those with p-value >0.01 were excluded. The remaining features were processed through stepwise backward elimination. In brief, in each elimination round, a batch of features were removed and performance evaluated. If the concordance index (c-index) decreased by <0.001, features were eliminated, otherwise a smaller subset was removed until the final set of features whose removal would cause performance degradation was found. These highly-contributing features were then used in the reduced models.

Both CPH and DeepSurv models were then trained using the final set of features, along with a model variant excluding all cholesterol variables and systolic blood pressure (substituted by heart rate for its digital feasibility), to enable calculation of risk for individuals who are not able to obtain these measurements.

**Comparison to other models**

Details of variable derivation for replication of Framingham risk score on the UKB dataset is summarised in *Supplementary Table 4*. Framingham score for each participant was calculated using the formulas for males and females published in the original article (8), and c-index was calculated using the predicted and actual time-to-CVD-event. Additionally, a CPH model was re-trained using the seven Framingham variables on the UKB dataset and compared to our findings. Sex-specific variants of our final models were trained to allow for closer comparison to these two scores.

**Statistical analysis**

In the summaries of cohort characteristics, participant numbers and percentages of total are shown for categorical and ordinal variables, whereas median and 1$^{st}$ and 3$^{rd}$ quartiles are shown for continuous variables. Statistical comparisons were performed using the Chi-squared test for categorical and ordinal variables and Kruskal-Wallis test for continuous variables.

Where detailed analysis of the results of CPH models is provided, hazard ratios (HR) with 95% confidence intervals (CIs), as well as the coefficients, are provided. P-values test the null hypothesis that the coefficient of each variable is equal to zero. Significance level was set to 0.05.

C-index was used as the metric for both models, with 95% CIs calculated using the percentile bootstrap resampling method (50 resampling rounds). Where detailed analysis of the results of CPH models is provided, log(HR) with 95% CIs are shown. P-values test the null hypothesis that the coefficient of each variable is equal to zero and significance level was set to 0.05. Calibration was evaluated at the 10-year time point using calibration plots and the Integrated Calibration Index (ICI), which is a mean weighted difference between observed and predicted probabilities, implemented in the Python lifelines library[2]. This article was written following the TRIPOD (Transparent Reporting of a Multivariable Prediction Model for Individual Prognosis or Diagnosis) guidelines, which are further elaborated in *Supplementary Table 5*.

# Results

**Population characteristics**

After exclusion of participants with pre-existing CVD, the study was conducted with 466,052 UKB participants. There were 42,377 participants (9.09%) who developed CVD during the observation period. The most common CVDs were chronic ischaemic heart disease (42.1%), myocardial infarction (20.3%), and stroke (15.4%). The full breakdown can be found in *Supplementary Table 2*. Median follow-up time was 11.5 years (IQR 10.7–12.3 years) and maximum follow-up time was 14.6 years.

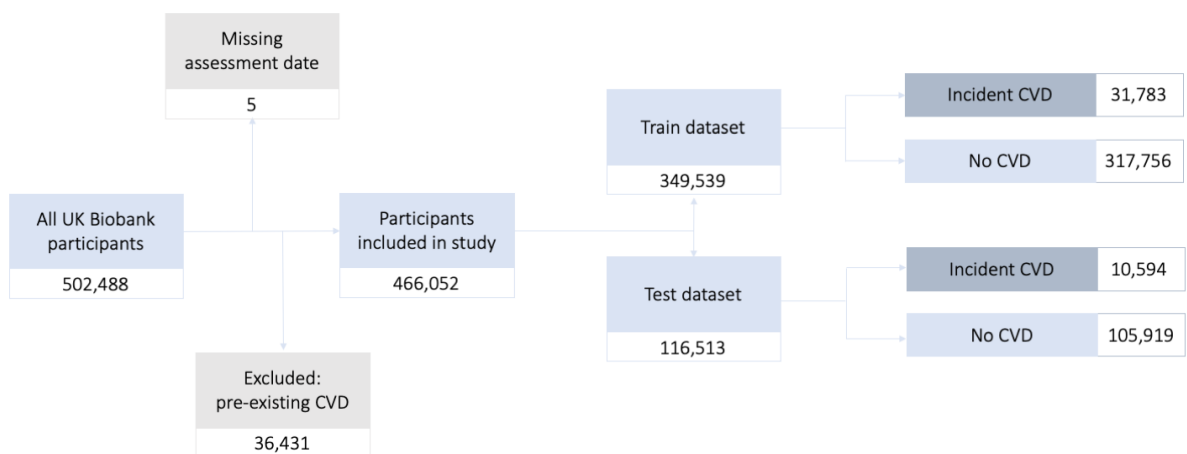

*Figure 1:* Numbers of participants included in the study with the breakdown of CVD incidence in the train and test datasets.

The studied dataset includes 44.1% men and 55.9% women (sex-specific demographic data shown in *Supplementary Table 6*), aged 37-73 at the time of recruitment (mean 56.2 ± 8.09). The participants were predominantly white, with a small proportion of other ethnic groups (2.2% Asian, 1.6% Black and 1.5% Other). Detailed demographic analysis for the variables selected in the final model can be found in *Table 2*.

|  | n (%) | | | P-Value (adjusted) |
|---|---|---|---|---|
|  | **All participants** | **No incident CVD** | **Incident CVD** |  |
| Total | 466052 | 423675 | 42377 |  |
| Sex: Female | 260673 (55.93) | 243390 (57.45) | 17283 (40.78) | <0.001 |
| Age, median [Q1,Q3] | 57 [50,63] | 57 [49,62] | 62 [56,66] | <0.001 |
| Has college or university degree, n (%) | 153353 (32.90) | 142831 (33.71) | 10522 (24.83) | <0.001 |
| Waist-to-hip ratio, median [Q1,Q3] | 0.87 [0.80,0.93] | 0.86 [0.80,0.93] | 0.91 [0.85,0.97] | <0.001 |
| Lost weight compared with 1 year ago, n (%) | 69005 (14.81) | 62106 (14.66) | 6899 (16.28) | <0.001 |
| Systolic blood pressure, median [Q1,Q3] | 137.50 [125.00,148.50] | 137.00 [124.50,147.50] | 142.00 [132.00,155.50] | <0.001 |
| Heart rate*, median [Q1,Q3] | 69.50 [62.50,75.50] | 69.50 [62.50,75.50] | 69.54 [62.50,77.00] | <0.001 |
| Total cholesterol, median [Q1,Q3] | 5.77 [5.06,6.41] | 5.77 [5.07,6.40] | 5.77 [4.97,6.45] | <0.001 |
| Cholesterol ratio, median [Q1,Q3] | 4.14 [3.43,4.66] | 4.14 [3.41,4.62] | 4.14 [3.66,4.98] | <0.001 |
| Currently does not smoke, n (%) | 416519 (89.37) | 380368 (89.78) | 36151 (85.31) | <0.001 |
| Smoked occasionally in the past, n (%) | 61282 (13.15) | 56096 (13.24) | 5186 (12.24) | <0.001 |
| Pack years of smoking, median [Q1,Q3] | 11.88 [0.00,22.29] | 11.00 [0.00,22.29] | 20.00 [0.00,22.29] | <0.001 |
| Never drinks alcohol, n (%) | 35946 (7.71) | 31707 (7.48) | 4239 (10.00) | <0.001 |
| Always adds salt to served food, n (%) | 22252 (4.77) | 19582 (4.62) | 2670 (6.30) | <0.001 |
| Hours spent outdoors in winter, median [Q1,Q3] | 1.93 [1.00,2.00] | 1.00 [1.00,2.00] | 1.93 [1.00,3.00] | <0.001 |
| Steady average usual walking pace, n (%) | 243706 (52.29) | 219976 (51.92) | 23730 (56.00) | <0.001 |
| Brisk usual walking pace, n (%) | 186297 (39.97) | 174169 (41.11) | 12128 (28.62) | <0.001 |
| Excellent self-rated health, n (%) | 80390 (17.25) | 75946 (17.93) | 4444 (10.49) | <0.001 |
| Good self-rated health, n (%) | 274677 (58.94) | 252295 (59.55) | 22382 (52.82) | <0.001 |
| Poor self-rated health, n (%) | 16742 (3.59) | 13477 (3.18) | 3265 (7.70) | <0.001 |
| Father diagnosed with heart disease, n (%) | 128063 (27.48) | 114859 (27.11) | 13204 (31.16) | <0.001 |
| Mother diagnosed with heart disease, n (%) | 81038 (17.39) | 71723 (16.93) | 9315 (21.98) | <0.001 |
| Sibling diagnosed with heart disease, n (%) | 34085 (7.31) | 28943 (6.83) | 5142 (12.13) | <0.001 |
| Diagnosis of depressive episode (F32), n (%) | 36410 (7.81) | 32466 (7.66) | 3944 (9.31) | <0.001 |
| Diagnosis of epilepsy (G40), n (%) | 4453 (0.96) | 3859 (0.91) | 594 (1.40) | <0.001 |
| Diagnosis of atrial fibrillation and flutter (I48), n (%) | 4925 (1.06) | 3371 (0.80) | 1554 (3.67) | <0.001 |
| Diagnosis of other cardiac arrhythmias (I49), n (%) | 3892 (0.84) | 3179 (0.75) | 713 (1.68) | <0.001 |
| Diagnosis of urinary tract infection or incontinence (I39), n (%) | 23218 (4.98) | 20629 (4.87) | 2589 (6.11) | <0.001 |
| Diagnosis of diabetes (E10, E11, E14), n (%) | 20054 (4.30) | 15919 (3.76) | 4135 (9.76) | <0.001 |
| Diagnosis of haematological cancer, n (%) | 1921 (0.41) | 1528 (0.36) | 393 (0.93) | <0.001 |
| Diagnosis of cellulitis (L03), n (%) | 3635 (0.78) | 3007 (0.71) | 628 (1.48) | <0.001 |
| Has diabetes-related eye disease, n (%) | 2602 (0.56) | 1860 (0.44) | 742 (1.75) | <0.001 |
| Fractured bones in the last 5 years, n (%) | 43598 (9.35) | 39197 (9.25) | 4401 (10.39) | <0.001 |
| Does not have any long-standing illness, disability or infirmity, n (%) | 318502 (68.34) | 295823 (69.82) | 22679 (53.52) | <0.001 |
| Number of operations, median [Q1,Q3] | 1 [1,2] | 1 [1,2] | 2 [1,3] | <0.001 |
| Regularly takes blood pressure medications, n (%) | 81084 (17.40) | 67673 (15.97) | 13411 (31.65) | <0.001 |
| Regularly takes insulin, n (%) | 4124 (0.88) | 3095 (0.73) | 1029 (2.43) | <0.001 |
| Regularly takes aspirin, n (%) | 44935 (9.64) | 37535 (8.86) | 7400 (17.46) | <0.001 |
| Regularly takes corticosteroids, n (%) | 3993 (0.86) | 3257 (0.77) | 736 (1.74) | <0.001 |
| Does not regularly take mineral supplements, fish oil or glucosamine, n (%) | 261167 (56.04) | 238037 (56.18) | 23130 (54.58) | <0.001 |
| Does not take insulin or medications for cholesterol/blood pressure, n (%) | 328612 (70.51) | 305358 (72.07) | 23254 (54.87) | <0.001 |
| Number of medications taken regularly, median [Q1,Q3] | 2 [0,3] | 1 [0,3] | 3 [1,5] | <0.001 |
| Experiences dyspnoea (R060), n (%) | 1729 (0.37) | 1334 (0.31) | 395 (0.93) | <0.001 |
| Experiences abdominal or pelvic pain (R10), n (%) | 21070 (4.52) | 18496 (4.37) | 2574 (6.07) | <0.001 |
| Experiences dizziness or giddiness (R42), n (%) | 1626 (0.35) | 1319 (0.31) | 307 (0.72) | <0.001 |
| Experiences syncope or collapse (R55), n (%) | 3354 (0.72) | 2732 (0.64) | 622 (1.47) | <0.001 |
| Has had wheeze or whistling in the chest in the last year, n (%) | 91158 (19.56) | 79415 (18.74) | 11743 (27.71) | <0.001 |
| Never feels any pain or discomfort in their chest, n (%) | 393719 (84.48) | 360851 (85.17) | 32868 (77.56) | <0.001 |

\* this variable used only as a substitute for systolic blood pressure in the model excluding cholesterol and systolic blood pressure

*Table 2: Summary of demographic characteristics of the studied cohort grouped by the outcomes. Last column shows p-value after comparing the incident CVD group with the non-CVD group. Comparisons were performed using the Chi-squared test for categories and Kruskal-Wallis test for continuous variables.*

**Baseline model performance and feature selection**

CPH and DeepSurv models were trained using the pre-processed dataset, containing 608 features. This baseline CPH model achieved a c-index of 0.7431 (95% CI 0.7422–0.7441) on the test dataset. DeepSurv, with optimised hyperparameters (***Supplementary Table 7***), achieved a c-index of 0.7461 (95% CI 0.7452–0.7469) on the test dataset. The CPH model was used for the subsequent feature selection from this baseline dataset.

Initially, 93 features were eliminated based on a p-value >0.01 in a univariate CPH model. The subsequent stepwise backward elimination excluded a further 465 features. The remaining 50 features were subjected to review by a clinician, further excluding three features (First Occurrence of ICD-10 codes F17 and H25, lamb/mutton intake 2-4 times a week).

**Performance of the reduced prediction models**

The remaining 47 features in the final reduced models include demographic measures (age, sex, holding a university degree), anthropometrics (waist-to-hip ratio), systolic blood pressure, total cholesterol, cholesterol ratio, range of pre-existing conditions, medications, symptoms, family history of heart disease, lifestyle measures (e.g. smoking, alcohol consumption), and self-rated health. A CPH model trained using these features achieved a c-index 0.7443 (95% CI 0.7441–0.7445) on the test dataset. A DeepSurv model showed concordance of 0.7446 (95% CI 0.7441–0.7452) on the test dataset.

The details of CPH feature coefficients and statistical analysis can be found in ***Figure 2*** and ***Supplementary Table 8***. Based on the p-value calculated in the CPH model, the top five risk factors include age, systolic blood pressure, diagnosis of atrial fibrillation and flutter (ICD-10 code I48), cholesterol ratio, and father with heart disease. The top five protective features include being a female, non-smoker, not experiencing any chest pain, brisk usual walking pace, and excellent self-rated health.

The mean predicted 10-year risk of developing CVD was 8.17% (95% CI 8.12–8.21), the mean observed risk was 7.96% (95% CI 7.92–8.01). The CPH models showed good calibration, with slight overestimation of the higher probabilities which were more sparsely represented in the dataset (***Figure 3***). The Integrated Calibration Index (ICI) was 0.295%.

To enable calculation of risk scores for individuals without access to cholesterol and blood pressure measurements, the models were re-trained after excluding total cholesterol and cholesterol ratio. Heart rate replaced systolic blood pressure as it is measurable via mobile device. The c-index of the CPH model decreased to 0.741 and c-index of the DeepSurv model decreased to 0.739 (***Table 3***).

|  | Before feature selection | After feature selection | Excluding cholesterol measurements + substituting BP |
|---|---|---|---|
| Number of features | 608 | 47 | 45 |
| CPH model C-index | 0.7431 [0.7422 – 0.7441] | 0.7443 [0.7441 – 0.7445] | 0.7409 [0.7407 – 0.7411] |
| DeepSurv model C-index | 0.7461 [0.7452 – 0.7469] | 0.7446 [0.7441 – 0.7452] | 0.7388 [0.7382 – 0.7393] |

***Table 3: C-index of trained models for test dataset predictions.*** *95% confidence interval is shown in square brackets. Cholesterol measurements include total cholesterol and cholesterol ratio, systolic blood pressure was substituted with heart rate. BP = blood pressure.*

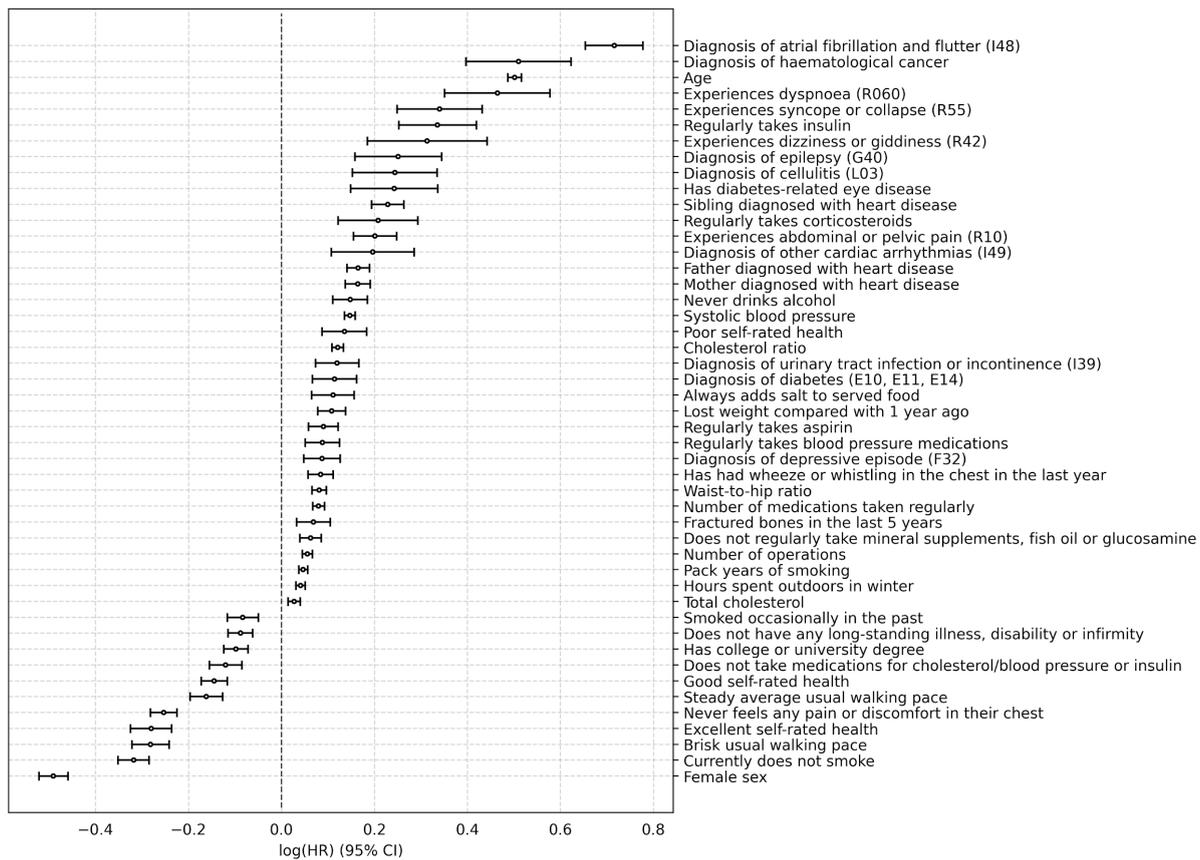

*Figure 2: Plot of Cox Proportional Hazards model coefficients for the general population.* Values show log(HR) ± 95% CI. HR = hazard ratio, CI = confidence interval.

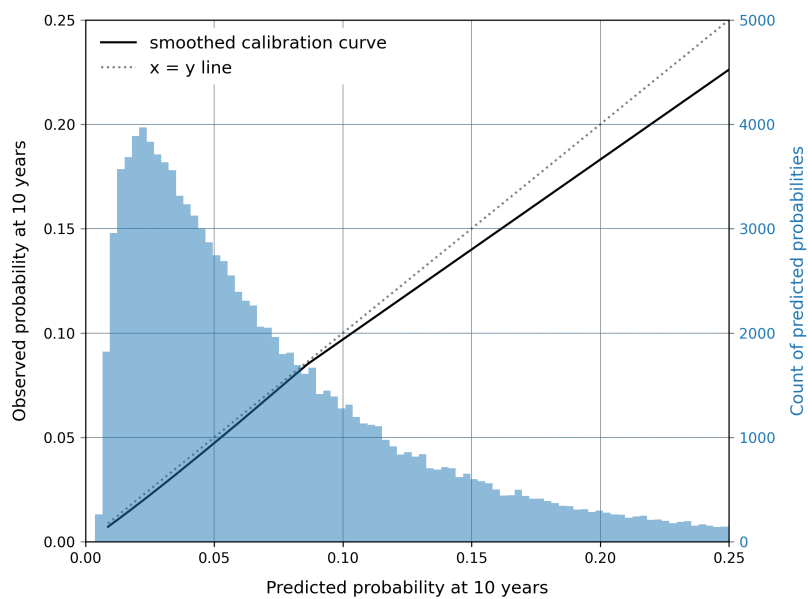

*Figure 3: Calibration plot for 10-year probabilities of developing CVD.* Results are shown for the final reduced Cox model evaluated on the test dataset. Smoothed calibration curve is shown along with the perfect calibration curve (x=y line). A histogram of the predicted probabilities of developing CVD for the participants in the test dataset is shown in blue.

**Comparison to existing risk scores**

Established risk scores, such as Framigham, provide separate models for males and females. To provide closer comparison to these scores, we trained sex-specific models using our final set of variables (excluding sex). The c-index of the male-specific model (both CPH and DeepSurv) was 0.72 (detailed results in *Supplementary Table 9*) and the female-specific model was 0.75 (detailed results in *Supplementary Table 10*).

Risk scores calculated using the Framingham formula achieved a c-index of 0.68 and 0.70 for male and females, respectively. The c-indices rose slightly after re-training the CPH model using the 7 Framingham variables on the UKB dataset. Comparison of all c-indices can be found in *Table 4*.

| Score | Method | C-index | | |
| --- | --- | --- | --- | --- |
| | | Men | Women | All participants |
| Framingham score | Risk scores calculated using published formula | 0.678 | 0.695 | 0.704 |
| | CPH model trained on UKB | 0.684 [0.684–0.685] | 0.714 [0.713–0.714] | 0.715 [0.715–0.715] |
| Our score | CPH model | 0.716 [0.716–0.717] | 0.748 [0.747–0.748] | 0.744 [0.744–0.744] |
| | DeepSurv model | 0.716 [0.715 – 0.717] | 0.747 [0.746–0.748] | 0.745 [0.744–0.745] |

*Table 4: Results of comparison of our model for Framingham score.* Test dataset c-indices shown. 95% confidence intervals are shown in the square brackets. UKB = UK Biobank.

## Discussion

Through our investigation of the UKB's sizable dataset, we were able to develop a model with a c-index of 0.745, showing very good predictive ability for CVD events over a 10-year period. The main features in the model are well-established across other risk scores: age, sex, hypertensive medication, systolic blood pressure, smoking status, and cholesterol. However, minimal differences were observed when cholesterol-related variables were excluded (0.744 vs. 0.741 using the Cox model), strengthening the case for use of our model in a remote setting and without mandatory cholesterol screening. The clinical demand for this exclusion is illustrated by the incorporation of a non-laboratory-based score along with the original Framingham CVD risk score (8). The observed modest decrease in c-index after exclusion of cholesterol may be explained by multicollinearity with other factors with high importance in our model: waist-to-height ratio, smoking status, family history, and hypertensive medication. Anthropometric measures, while traditionally viewed as valued parameters only of obesity and diabetes (32), are gaining traction in risk modelling across a wider range of non-communicable and communicable diseases alike (33). The expanding impact of the digital revolution on healthcare makes these features especially relevant owing to novel, accessible technologies which can capture a broad range of anthropometric information using only a smartphone camera (34).

Alcohol consumption and smoking habits, while traditional factors, present in an interesting manner. For the former, as categorical UKB fields were one-hot encoded, only alcohol abstinence passed feature selection, with other degrees of consumption not significant. The detrimental predictive significance of alcohol abstinence can be attributed to the average poorer health status and higher prevalence of chronic conditions and neurological problems of never- and former-drinkers, when compared to ever-drinkers (35). For tobacco, as expected, currently not smoking is protective and increasing pack years confers increased risk of CVD. However, having smoked occasionally was unexpectedly calculated as protective, likely as this group is enriched for people who have since ceased smoking, having never smoked regularly.

In addition to traditional risk factors and anthropometric variables, several other variables not present in any conventional risk model have shown to be significant contributors to our model. Notably, the inclusion of reported symptoms in our analysis sets the resultant model apart from others. The symptoms of dyspnoea, syncope, dizziness, wheezing, and chest pain are recognised as clinical indicators of possible cardiac disease. Our model supports their potential to highlight subclinical, underlying cardiovascular pathology. Similarly, abdominal and pelvic pain may represent undiagnosed comorbidity associated with our CVD outcomes, including inflammatory bowel disease, endometriosis, or aortic aneurysm (36,37). The extent of one's education has also emerged as an important contributing social factor in our model, reinforcing the findings of recent studies which show education as a significant predictor of CVD acquisition and mortality (38).

Furthermore, several diagnoses not traditionally associated with the incidence of CVD were found to be significant contributors to the CPH model. A diagnosis of urinary tract infections (UTI) or incontinence, as well as epilepsy, are significant contributors independent of all other features. UTI and incontinence may be explained by the raised cardiovascular risk of the proinflammatory state, but warrants further investigation. Epidemiological studies have shown a raised CVD risk profile in those with epilepsy, but this is mostly attributed to concomitant risk factors (39).

Self-rated health and prior diagnosis of a depressive episode, both associated with increased CVD risk (40,41), are also significant features, stressing the assessment of psychological aspects of health when calculating a risk for CVDs. To a smaller, but still important extent, two physical activity-related behaviours—walking pace and hours spent outdoors (in winter)—have contributed to the performance of the model, both known to influence the risk of CVD-related events (42,43). Lastly, a family history of heart disease showed strong predictive power, with sibling heart disease ranking significantly higher than parental. This is likely as a sibling diagnosis has been caused both by similar genetic and environmental factors, whereas parental diagnoses reflect pertinent environmental contributors to a lesser extent. Despite its well-known contribution to CVD occurrence, the QRISK3 is the only risk model that currently incorporates family history in the calculation, but our results show that more granular coverage of family history could lead to better risk assessment (7).

By using CPH and DeepSurv to develop risk scores, we compared an established statistical method with a more novel machine learning approach to survival analysis. We hypothesized that use of DeepSurv might result in superior predictions by identifying complex interactions between variables. However, performances of both models were similar, implying a minor contribution of non-linear interactions within the large feature space.

The minimal performance drop when excluding the less accessible features of cholesterol and blood pressure, together with the fact that many features are modifiable, supports the feasibility of feature collection through a digital application, in a remote setting, and on a regular basis. This enables more engaging risk management and eliminates the need for interaction with a healthcare provider. Our 'digitally-collected' score slightly overperformed Framingham risk score (c-index 0.745 vs 0.704).

Due to increasing ownership, mobile phones and smart wearables have become a priority medium for digital health interventions. The use of digitally-collected features to calculate CVD risk will allow patient-facing solutions, transferring responsibility of health status and lifestyle choices to patients rather than clinical teams. Furthermore, gamification and goal setting has been shown to increase the success of lifestyle change interventions (44). Several risk variables, particularly those related to diet (salt added to food, nutritional supplement intake and alcohol), as well as activity (hours spent outdoors in winter) are adjustable in the short term, giving direct feedback to users. By having a patient-facing digital risk score that is updated regularly, patients with high CVD risk may find increased motivation from the positive feedback of seeing their CVD risk change with improvements in lifestyle. The assessment of cardiovascular risk is a mainstay of clinical guidelines (5), and the ability to calculate this remotely with very limited face-to-face clinical resources could have significant beneficial cost and patient accessibility implications.

A notable limitation of our study stems from the UKB cohort being unrepresentative of the general population across various domains (45). First, ethnic diversity is very low with 94% of all participants identifying as White and, second, age at recruitment was restricted to 37-73. Our analysis, therefore, was not able to account for differences in CVD risk across ethnicities, which is a significant known factor for CVD incidence, nor in younger age groups, thus use of our model in these populations should be interpreted with some reserve. Importantly, the restricted age distribution also represents a higher risk population, implying validation of the model in a more representative sample would result in a higher performance, as seen when QRISK3 is calculated for UKB participants (representative population: 0.88; UKB: 0.76) (7,46). Last, the UKB cohort is reported to be healthier and wealthier than the general population (45). This is a significant factor when genetic aspects account for only ~32% of coronary artery disease occurrence (47), and thereby may result in an underappreciation of the influence of detrimental lifestyle choices. Future work will concern external validation of the model which will allow conclusions to be drawn about its use across geographies and populations.

While our aim was to limit model bias by applying a data-driven approach to feature selection and excluding features with a large proportion of missing data, choosing CPH model-based feature selection may have biased the final features in the reduced model. It is likely that this has also limited the performance of the DeepSurv model, by selecting features which specifically enhanced the performance of the CPH model.

In conclusion, DiCAVA, our 10-year CVD risk model, has very good predictive capacity and contains significant predictors not previously described by existing risk scores in the literature. We demonstrated its feasible utility in a remote setting where cholesterol and blood pressure measurements may not always be convenient, highlighting that even the most established predictors are not always essential.

## Acknowledgements

The authors would like to thank Sam Nikbakhtian and Michele Colombo for their contributions.

## Funding

This research was funded by Huma Therapeutics Ltd. The funders had no role in study design, data collection and analysis, decision to publish, or preparation of the manuscript.

## Disclosure statement

N.D., A.B.R., A.D., B.D.O., D.M., M.A., and D.P. are employees of Huma Therapeutics Ltd.

# Supplementary Material

*Supplementary Table 1: Main cardiovascular risk scores used in clinical practice*

| Score name | Variables included |
|---|---|
| Framingham (8) | Sex, age, total cholesterol, HDL cholesterol, systolic blood pressure, smoking status, diabetes mellitus, hypertensive treatment |
| SCORE (10) | Sex, age, total cholesterol or total cholesterol to HDL cholesterol ratio, systolic blood pressure, smoking status. Versions for use in high and low-risk countries |
| ASSIGN-SCORE (48) | Sex, age, total cholesterol, HDL cholesterol, systolic blood pressure, smoking – number of cigarettes, diabetes mellitus, area-based index of deprivation, family history |
| QRISK3 (7) | Sex, age, total cholesterol to HDL cholesterol ratio, systolic blood pressure, smoking status, diabetes mellitus, area-based index of deprivation, family history, BMI, blood pressure treatment, ethnicity and chronic diseases |
| Pooled Studies Cohort Equation (6) | Age, sex, race (white or other/African American), total cholesterol, HDL cholesterol, systolic blood pressure, blood pressure treatment, diabetes mellitus, smoking |
| PROCAM (49) | Age, sex, LDL-cholesterol, HDL-cholesterol, diabetes mellitus, smoking, systolic blood pressure |
| CUORE (50) | Age, sex, systolic blood pressure, total cholesterol, HDL cholesterol, blood pressure treatment and smoking habits |
| Globorisk (9) | Age, sex, smoking, total cholesterol, diabetes mellitus, systolic blood pressure |

Adapted and modified from: *Piepoli, Massimo F et al. "2016 European Guidelines on cardiovascular disease prevention in clinical practice: The Sixth Joint Task Force of the European Society of Cardiology and Other Societies on Cardiovascular Disease Prevention in Clinical Practice (constituted by representatives of 10 societies and by invited experts)Developed with the special contribution of the European Association for Cardiovascular Prevention & Rehabilitation (EACPR)." European heart journal vol. 37,29 (2016): 2315-2381.* [doi:10.1093/eurheartj/ehw106](doi:10.1093/eurheartj/ehw106)

***Supplementary Table 2: Overview of UK Biobank fields used for the definition of CVD outcome.*** *List of UKB fields and breakdown of count for the CVD types among the outcomes are shown.*

| CVD type | Total count | UK Biobank field number | Description |
|---|---|---|---|
| Heart attack | 8609 | 42000 | Date of myocardial infarction |
| | | 131298 | Date I21 first reported (acute myocardial infarction) |
| | | 131300 | Date I22 first reported (subsequent myocardial infarction) |
| | | 131302 | Date I23 first reported (certain current complications following acute myocardial infarction) |
| | | 131304 | Date I24 first reported (other acute ischaemic heart diseases) |
| Heart failure | 5736 | 131354 | Date I50 first reported (heart failure) |
| | | 41270 & 41280 | I110 (Hypertensive heart disease with (congestive) heart failure I130 (Hypertensive heart and renal disease with (congestive) heart failure) I132 (Hypertensive heart and renal disease with both (congestive) heart failure and renal failure) |
| Chronic ischaemic heart disease | 17963 | 131296 | Date I20 first reported (angina pectoris) |
| | | 131306 | Date I25 first reported (chronic ischaemic heart disease) |
| Stroke | 6525 | 42006 | Date of stroke |
| | | 131360 | Date I60 first reported (subarachnoid haemorrhage) |
| | | 131362 | Date I61 first reported (intracerebral haemorrhage) |
| | | 131366 | Date I63 first reported (cerebral infarction) |
| | | 131368 | Date I64 first reported (stroke, not specified as haemorrhage or infarction) |
| | | 131058 | Date G46 first reported (vascular syndromes of brain in cerebrovascular disease) |
| Transient ischaemic attack | 3544 | 131056 | Date G45 first reported (transient cerebral ischaemic attacks and related syndromes) |
| **TOTAL** | **42377** | - | - |

***Supplementary Table 3: Summary of the UK Biobank fields used in this study.*** *UKB field numbers and readable names are shown. Total numbers after processing the features (including feature engineering, one-hot encoding and rare feature exclusions) are shown.*

| Category | Number of features | List of UKB fields/ICD-10 codes |
|---|---|---|
| Demographics | 14* | 31 (sex), 21022 (age), 6138 (qualifications), 6142 (current employment status) |
| Anthropometrics | 6* | 48 (waist circumference), 49 (hip circumference), 58 (standing height), 21002 (weight), 21001 (BMI) |
| Biological measurements | 6* | 102 (heart rate), 4080 (systolic blood pressure), 4090 (diastolic blood pressure), 30690 (cholesterol), 30760 (HDL cholesterol) |
| Environmental measurements | 8 | 24012 (distance to nearest major road), 24003 (NO2 air pollution), 24004 (NO air pollution), 24005 (PM2.5 air pollution), 24006 (PM10 air pollution), 24011 (traffic intensity on the nearest major road), 24015 (length of major roads within 100m), 24506 (natural environment percentage) |
| Activity measurements | 12* | 1050 (time spent outdoors in summer), 1060 (time spent outdoors in winter), 1070 (time spent watching TV), 1080 (time spent using computer), 1090 (time spent driving), 22037 (minutes per week of walking), 22038 (minutes per week of moderate activity), 22039 (minutes per week of vigorous activity), 924 (usual walking pace) |
| Sleep habits | 20* | 1160 (sleep duration), 1200 (insomnia), 1220 (daytime sleeping), 1180 (chronotype), 1210 (snoring), 1190 (napping during day), 1170 (getting up in the morning) |
| Alcohol habits | 7* | 1558 (alcohol intake frequency), 1568 (red wine intake), 1578 (champagne and white wine intake), 1588 (beer and cider intake), 1598 (spirits intake), 5364 (intake of other alcoholic drinks) |
| Smoking habits | 13* | 20116 (smoking status), 1239 (current smoking), 1249 (past smoking), 3456 (number of cigarettes per day), 20161 (pack years of smoking), 20160 (ever smoked) |
| Diet | 41* | 1349 (processed meat intake), 1289 (cooked vegetable intake), 1299 (raw vegetable intake), 1309 (fresh fruit intake), 1359 (poultry intake), 1369 (beef intake), 1379 (lamb/mutton intake), 1389 (pork intake), 1408 (cheese intake), 1478 (salt added to served food), 1488 (tea intake), 1498 (coffee intake), 1528 (water intake) |
| Family history | 34* | 20107 (illnesses of father), 20110 (illnesses of mother), 20111 (illnesses of sibling) |
| Medications | 35* | 6177 & 6153 (medications for cholesterol, blood pressure or diabetes), 6153 (medications for pain, constipation or heartburn), 6179 (mineral and other supplements), 6155 (vitamin and mineral supplements), 137 (number of medications taken), 20003 (treatment/medications codes)** |
| Other health-related fields | 50* | 2296 (falls in the last year), 2463 (fractured bones in the last 5 years), 2188 (long-standing illness), 2306 (weight change compared to a year ago), 136 (number of operations), 2415 (had major operations), 6159 (pain in the last month), 2247 (hearing problems), 4803 (tinnitus), 6149 (mouth and dental problems), 6148 (eye problems), 2316 (wheeze or whistling in the chest in the last year), 2335 (chest pain or discomfort), 2178 (self-rated health) |
| First occurrences of ICD-10 diagnoses | 265*** | All First Occurrences fields except those included in outcomes<br>Notes:<br>- E10–E14 combined into Diabetes |
| ICD-10 codes | 97*** | 41270 & 41280 (codes 'L02', 'L03', 'L40', 'L72', 'L82', 'L90', 'L98', 'O02', 'O03', 'O04', 'O20', 'O26', 'O32', 'O34', 'O36', 'O42', 'O47', 'O48', 'O63', 'O68', 'O70', 'O72', 'O80', 'O99', 'R00', 'R04', 'R05', 'R060', 'R10', 'R11', 'R12', 'R13', 'R20', 'R22', 'R31', 'R32', 'R33', |

| | | 'R35', 'R39', 'R42', 'R50', 'R51', 'R55', 'R59', 'R63', 'R69', 'R79', 'R87', 'R93', 'R94', 'S01', 'S02', 'S09', 'S22', 'S42', 'S52', 'S61', 'S62', 'S72', 'S82', 'W01', 'W10', 'W19', 'Z04', 'Z08', 'Z09', 'Z12', 'Z13', 'Z30', 'Z34', 'Z35', 'Z36', 'Z37', 'Z42', 'Z43', 'Z45', 'Z46', 'Z47', 'Z48', 'Z50', 'Z60', 'Z80', 'Z85', 'Z86', 'Z87', 'Z88', 'Z90', 'Z91', 'Z921', 'Z922', 'Z93', 'Z96', 'Z98') |
|---|---|---|
| | | Notes:<br>- C81–C96 & D45–D47 combined into Haematological cancers<br>- D37.5 & C19–C21 & D12 & C39 & C44 & C78.0–C78.3 & Z85.2–Z85.3 combined into Non-hematological cancers and split into three groups based on time of diagnosis (less than 1 year, 1-5 years or more than 5 years before assessment) |
| **Total** | 608 | - |

*after processing (one-hot encoding, feature engineering, exclusion of rare categories)
**clinician-curated lists of corticosteroids, migraine medications, atypical antipsychotics, erectile dysfunction medications, mental health illness medications and hypertension medications, details shown below
***features with occurrence less than 0.2% in the population were excluded

**Corticosteroids**: 1140874790, 1140874816, 1140874822, 1140874896, 1140874930, 1140874936, 1140874940, 1140874944, 1140874950, 1140874954, 1140874956, 1140874976, 1140874978, 1140910424, 1140910634
**Migraine medications:** 1141151284, 1141151288, 1141157332, 1140884412, 1140911658, 1141150620, 1141150624, 1141167932, 1141167940, 1141172728, 1141172628, 1141185436, 1141185448, 1141192666, 1141192670
**Atypical antipsychotics:** 1140867420, 1140867432, 1140867444, 1140882320, 1140928916, 1141167976, 1141152848, 1141152860, 1141177762, 1140927956, 1140927970, 1141169714, 1141169722, 1141195974, 1141202024
**Erectile dysfunction medications:** 1141168936, 1141168944, 1141168946, 1141168948, 1141187810, 1141187814, 1141187818, 1141192248, 1141192256, 1141192258, 1141192260
**Mental health illness medications:** 1140879616, 1140921600, 1140879540, 1140867878, 1140916282, 1140909806, 1140867888, 1141152732, 1141180212, 1140879634, 1140867876, 1140882236, 1141190158, 1141200564, 1140867726, 1140879620, 1140867818, 1140879630, 1140879628, 1141151946, 1140867948, 1140867624, 1140867756, 1140867884, 1141151978, 1141152736, 1141201834, 1140867690, 1140867640, 1140867920, 1140867850, 1140879544, 1141200570, 1140867934, 1140867758, 1140867914, 1140867820, 1141151982, 1140882244, 1140879556, 1140867852, 1140867860, 1140917460, 1140867938, 1140867856, 1140867922, 1140910820, 1140882312, 1140867944, 1140867784, 1140867812, 1140867668, 1140928916, 1141152848, 1140867444, 1140879658, 1140868120, 1141153490, 1140867304, 1141152860, 1140867168, 1141195974, 1140867244, 1140867152, 1140909800, 1140867420, 1140879746, 1141177762, 1140867456, 1140867952, 1140867150, 1141167976, 1140882100, 1140867342, 1140863416, 1141202024, 1140882098, 1140867184, 1140867092, 1140882320, 1140910358, 1140867208, 1140909802, 1140867134, 1140867306, 1140867210, 1140867398, 1140867078, 1140867218, 1141201792, 1141200458, 1140867136, 1140879750, 1140867180, 1140867546, 1140927956, 1140867490, 1140867494, 1140867498, 1140867500, 1140867504, 1140867518, 1140867520
**Hypertension medications:** 1140866144, 1140866330, 1141146124, 1141146126, 1141146128, 1140860308, 1140864202, 1140866146, 1140851364, 1140860332, 1140860404, 1140860422, 1140860562, 1140860738, 1140860764, 1140860790, 1140864950, 1140864952, 1140866138, 1140866162, 1140926778, 1141151016, 1141172682, 1141187788, 1141201038, 1141188636, 1140851362, 1140851660, 1140866164, 1141180592, 1140866078, 1140866092, 1140866094, 1140866096, 1140866102, 1140866104, 1140866108, 1140866110, 1141152998

***Supplementary Table 4: Summary of the variables used for replication of Framingham risk score.***
*Participants with any missing values in these variables were excluded from the analysis.*

| Framingham score variable | UKBB fields used | Processing steps |
|---|---|---|
| Age | 31 | None |
| Total cholesterol | 30690 | Converted from mmol/L to mg/dL (factor 38.67) |
| HDL cholesterol | 30760 | Converted from mmol/L to mg/dL (factor 38.67) |
| Systolic blood pressure treated | 4080<br>6177 (value 2 only)<br>6153 (value 2 only)<br>20003 (curated list medications*) | Mean of the two blood pressure measurements taken, split into two columns based on whether they take hypertensive medicines or not |
| Systolic blood pressure non-treated | | |
| Current smoker | 20116 | Converted into a binary field for participants who answered they are current smokers (value 2) |
| Diabetes | 130706<br>130708<br>130710<br>130712<br>130714 | Censored all diagnoses after assessment date, combined all columns into one as a binary variable |
| Outcome | Same as our model | |

* 1140866144, 1140866330, 1141146124, 1141146126, 1141146128, 1140860308, 1140864202, 1140866146, 1140851364, 1140860332, 1140860404, 1140860422, 1140860562, 1140860738, 1140860764, 1140860790, 1140864950, 1140864952, 1140866138, 1140866162, 1140926778, 1141151016, 1141172682, 1141187788, 1141201038, 1141188636, 1140851362, 1140851660, 1140866164, 1141180592, 1140866078, 1140866092, 1140866094, 1140866096, 1140866102, 1140866104, 1140866108, 1140866110, 1141152998

*Supplementary Table 5: TRIPOD Guidelines Report.*

| Section/Topic | Item | | Checklist Item | Page |
|---|---|---|---|---|
| **Title and abstract** | | | | |
| Title | 1 | D;V | Identify the study as developing and/or validating a multivariable prediction model, the target population, and the outcome to be predicted. | 1 |
| Abstract | 2 | D;V | Provide a summary of objectives, study design, setting, participants, sample size, predictors, outcome, statistical analysis, results, and conclusions. | 2 |
| **Introduction** | | | | |
| Background and objectives | 3a | D;V | Explain the medical context (including whether diagnostic or prognostic) and rationale for developing or validating the multivariable prediction model, including references to existing models. | 3 |
| | 3b | D;V | Specify the objectives, including whether the study describes the development or validation of the model or both. | 3 |
| **Methods** | | | | |
| Source of data | 4a | D;V | Describe the study design or source of data (e.g., randomized trial, cohort, or registry data), separately for the development and validation data sets, if applicable. | 4 |
| | 4b | D;V | Specify the key study dates, including start of accrual; end of accrual; and, if applicable, end of follow-up. | 4 |
| Participants | 5a | D;V | Specify key elements of the study setting (e.g., primary care, secondary care, general population) including number and location of centres. | 4 |
| | 5b | D;V | Describe eligibility criteria for participants. | 4 |
| | 5c | D;V | Give details of treatments received, if relevant. | N/A |
| Outcome | 6a | D;V | Clearly define the outcome that is predicted by the prediction model, including how and when assessed. | 4, Suppl. Table 2 |
| | 6b | D;V | Report any actions to blind assessment of the outcome to be predicted. | N/A |
| Predictors | 7a | D;V | Clearly define all predictors used in developing or validating the multivariable prediction model, including how and when they were measured. | 4, Suppl. Table 3 |
| | 7b | D;V | Report any actions to blind assessment of predictors for the outcome and other predictors. | N/A |
| Sample size | 8 | D;V | Explain how the study size was arrived at. | 4, Fig. 1 |
| Missing data | 9 | D;V | Describe how missing data were handled (e.g., complete-case analysis, single imputation, multiple imputation) with details of any imputation method. | 4 |
| Statistical analysis methods | 10a | D | Describe how predictors were handled in the analyses. | 4 |
| | 10b | D | Specify type of model, all model-building procedures (including any predictor selection), and method for internal validation. | 4–5 |
| | 10c | V | For validation, describe how the predictions were calculated. | N/A |
| | 10d | D;V | Specify all measures used to assess model performance and, if relevant, to compare multiple models. | 5 |
| | 10e | V | Describe any model updating (e.g., recalibration) arising from the validation, if done. | N/A |
| Risk groups | 11 | D;V | Provide details on how risk groups were created, if done. | N/A |
| Development vs. validation | 12 | V | For validation, identify any differences from the development data in setting, eligibility criteria, outcome, and predictors. | N/A |
| **Results** | | | | |
| Participants | 13a | D;V | Describe the flow of participants through the study, including the number of participants with and without the outcome and, if applicable, a summary of the follow-up time. A diagram may be helpful. | 6, Fig. 1 |
| | 13b | D;V | Describe the characteristics of the participants (basic demographics, clinical features, available predictors), including the number of participants with missing data for predictors and outcome. | 6, Table 2 |
| | 13c | V | For validation, show a comparison with the development data of the distribution of important variables (demographics, predictors and outcome). | N/A |
| Model development | 14a | D | Specify the number of participants and outcome events in each analysis. | Fig. 1 |
| | 14b | D | If done, report the unadjusted association between each candidate predictor and outcome. | N/A |
| Model specification | 15a | D | Present the full prediction model to allow predictions for individuals (i.e., all regression coefficients, and model intercept or baseline survival at a given time point). | 8–9, Fig. 2, Suppl. Table 8 |
| | 15b | D | Explain how to use the prediction model. | 12 |
| Model performance | 16 | D;V | Report performance measures (with CIs) for the prediction model. | 8, Table 3 |
| Model-updating | 17 | V | If done, report the results from any model updating (i.e., model specification, model performance). | N/A |
| **Discussion** | | | | |
| Limitations | 18 | D;V | Discuss any limitations of the study (such as non-representative sample, few events per predictor, missing data). | 12 |
| Interpretation | 19a | V | For validation, discuss the results with reference to performance in the development data, and any other validation data. | N/A |
| | 19b | D;V | Give an overall interpretation of the results, considering objectives, limitations, results from similar studies, and other relevant evidence. | 11-12 |
| Implications | 20 | D;V | Discuss the potential clinical use of the model and implications for future research. | 11–12 |
| **Other information** | | | | |
| Supplementary information | 21 | D;V | Provide information about the availability of supplementary resources, such as study protocol, Web calculator, and data sets. | Supp. material |
| Funding | 22 | D;V | Give the source of funding and the role of the funders for the present study. | Funding statement |

*Supplementary Table 6: Summary of demographic characteristics of the studied cohort grouped by the sex.*

|  | n (%) | | |
| --- | --- | --- | --- |
|  | **All participants** | **Males** | **Females** |
| Total | 466052 | 205379 | 260673 |
| Incident CVD, n (%) | 42377 (9.09) | 25094 (12.22) | 17283 (6.63) |
| Age, median [Q1,Q3] | 57.00 [50.00,63.00] | 57.00 [49.00,63.00] | 57.00 [50.00,63.00] |
| Has college or university degree, n (%) | 153353 (32.90) | 71244 (34.69) | 82109 (31.50) |
| Waist-to-hip ratio, median [Q1,Q3] | 0.87 [0.80,0.93] | 0.93 [0.89,0.97] | 0.81 [0.77,0.86] |
| Lost weight compared with 1 year ago, n (%) | 69005 (14.81) | 29200 (14.22) | 39805 (15.27) |
| Systolic blood pressure, median [Q1,Q3] | 137.50 [125.00,148.50] | 138.00 [129.50,150.50] | 134.50 [122.00,146.00] |
| Total cholesterol, median [Q1,Q3] | 5.77 [5.06,6.41] | 5.65 [4.90,6.23] | 5.77 [5.19,6.54] |
| Cholesterol ratio, median [Q1,Q3] | 4.14 [3.43,4.66] | 4.17 [3.78,5.09] | 3.93 [3.22,4.26] |
| Currently does not smoke, n (%) | 416519 (89.37) | 179324 (87.31) | 237195 (90.99) |
| Smoked occasionally in the past, n (%) | 61282 (13.15) | 26922 (13.11) | 34360 (13.18) |
| Pack years of smoking, median [Q1,Q3] | 11.88 [0.00,22.29] | 18.00 [0.00,22.29] | 7.00 [0.00,22.29] |
| Never drinks alcohol, n (%) | 35946 (7.71) | 12201 (5.94) | 23745 (9.11) |
| Always adds salt to served food, n (%) | 22252 (4.77) | 10444 (5.09) | 11808 (4.53) |
| Hours spent outdoors in winter, median [Q1,Q3] | 1.93 [1.00,2.00] | 2.00 [1.00,3.00] | 1.00 [1.00,2.00] |
| Steady average usual walking pace, n (%) | 243706 (52.29) | 106666 (51.94) | 137040 (52.57) |
| Brisk usual walking pace, n (%) | 186297 (39.97) | 84050 (40.92) | 102247 (39.22) |
| Excellent self-rated health, n (%) | 80390 (17.25) | 34499 (16.80) | 45891 (17.60) |
| Good self-rated health, n (%) | 274677 (58.94) | 118208 (57.56) | 156469 (60.03) |
| Poor self-rated health, n (%) | 16742 (3.59) | 7988 (3.89) | 8754 (3.36) |
| Father diagnosed with heart disease, n (%) | 128063 (27.48) | 52736 (25.68) | 75327 (28.90) |
| Mother diagnosed with heart disease, n (%) | 81038 (17.39) | 29927 (14.57) | 51111 (19.61) |
| Sibling diagnosed with heart disease, n (%) | 34085 (7.31) | 12943 (6.30) | 21142 (8.11) |
| Diagnosis of depressive episode (F32), n (%) | 36410 (7.81) | 11848 (5.77) | 24562 (9.42) |
| Diagnosis of epilepsy (G40), n (%) | 4453 (0.96) | 2111 (1.03) | 2342 (0.90) |
| Diagnosis of atrial fibrillation and flutter (I48), n (%) | 4925 (1.06) | 3229 (1.57) | 1696 (0.65) |
| Diagnosis of other cardiac arrhythmias (I49), n (%) | 3892 (0.84) | 1770 (0.86) | 2122 (0.81) |
| Diagnosis of urinary tract infection or incontinence (I39), n (%) | 23218 (4.98) | 4317 (2.10) | 18901 (7.25) |
| Diagnosis of diabetes (E10, E11, E14), n (%) | 20054 (4.30) | 11900 (5.79) | 8154 (3.13) |
| Diagnosis of haematological cancer, n (%) | 1921 (0.41) | 1070 (0.52) | 851 (0.33) |
| Diagnosis of cellulitis (L03), n (%) | 3635 (0.78) | 2076 (1.01) | 1559 (0.60) |
| Has diabetes-related eye disease, n (%) | 2602 (0.56) | 1578 (0.77) | 1024 (0.39) |
| Fractured bones in the last 5 years, n (%) | 43598 (9.35) | 17567 (8.55) | 26031 (9.99) |
| Does not have any long-standing illness, disability or infirmity, n (%) | 318502 (68.34) | 137219 (66.81) | 181283 (69.54) |
| Number of operations, median [Q1,Q3] | 1.00 [1.00,2.00] | 1.00 [0.00,2.00] | 2.00 [1.00,3.00] |
| Regularly takes blood pressure medications, n (%) | 81084 (17.40) | 40223 (19.58) | 40861 (15.68) |
| Regularly takes insulin, n (%) | 4124 (0.88) | 2339 (1.14) | 1785 (0.68) |
| Regularly takes aspirin, n (%) | 44935 (9.64) | 25435 (12.38) | 19500 (7.48) |
| Regularly takes corticosteroids, n (%) | 3993 (0.86) | 1689 (0.82) | 2304 (0.88) |
| Does not regularly take mineral supplements, fish oil or glucosamine, n (%) | 261167 (56.04) | 126147 (61.42) | 135020 (51.80) |
| Does not take medications for cholesterol/blood pressure or insulin, n (%) | 328612 (70.51) | 146639 (71.40) | 181973 (69.81) |
| Number of medications taken regularly, median [Q1,Q3] | 2.00 [0.00,3.00] | 1.00 [0.00,3.00] | 2.00 [1.00,4.00] |
| Experiences dyspnoea (R060), n (%) | 1729 (0.37) | 682 (0.33) | 1047 (0.40) |
| Experiences abdominal or pelvic pain (R10), n (%) | 21070 (4.52) | 6138 (2.99) | 14932 (5.73) |
| Experiences dizziness or giddiness (R42), n (%) | 1626 (0.35) | 658 (0.32) | 968 (0.37) |
| Experiences syncope or collapse (R55), n (%) | 3354 (0.72) | 1691 (0.82) | 1663 (0.64) |
| Has had wheeze or whistling in the chest in the last year, n (%) | 91158 (19.56) | 42605 (20.74) | 48553 (18.63) |
| Never feels any pain or discomfort in their chest, n (%) | 393719 (84.48) | 171047 (83.28) | 222672 (85.42) |

*Supplementary Table 7: Summary of the best-performing DeepSurv models.*

| Hyperparameter | Baseline model | Reduced model |
| --- | --- | --- |
| Activation | ReLU | ReLU |
| Batch normalisation | Yes | Yes |
| Dropout | 0.6172 | 0.3338 |
| Weight decay | 0.0261 | 0.0596 |
| Learning rate | 0.000130 | 0.000309 |
| Optimiser | Adam | Adam |
| Hidden layers shape | 256 | 32 x 32 |

***Supplementary Table 8: Summary of the Cox Proportional Hazards model for all participants.*** *The table displays coefficients = log(HR), hazard ratios and 95% confidence intervals, as well as -log2 (p-value), where p < 0.05 an null hypothesis states that the coefficient is equal to 0.*

| Covariate | log(HR) | CI log(HR) lower 95% | CI log(HR) upper 95% | -log2 (p-value) |
|---|---|---|---|---|
| Diagnosis of atrial fibrillation and flutter (I48) | 0.716 | 0.654 | 0.778 | 375.462 |
| Diagnosis of haematological cancer | 0.51 | 0.397 | 0.623 | 60.115 |
| Age | 0.501 | 0.487 | 0.516 | inf |
| Experiences dyspnoea (R060) | 0.464 | 0.351 | 0.578 | 49.667 |
| Experiences syncope or collapse (R55) | 0.34 | 0.249 | 0.432 | 41.594 |
| Regularly takes insulin | 0.336 | 0.252 | 0.419 | 48.462 |
| Experiences dizziness or giddiness (R42) | 0.313 | 0.184 | 0.442 | 19.031 |
| Diagnosis of epilepsy (G40) | 0.251 | 0.158 | 0.344 | 22.814 |
| Diagnosis of cellulitis (L03) | 0.243 | 0.153 | 0.334 | 22.66 |
| Has diabetes-related eye disease | 0.242 | 0.148 | 0.336 | 21.131 |
| Sibling diagnosed with heart disease | 0.229 | 0.194 | 0.263 | 126.759 |
| Regularly takes corticosteroids | 0.207 | 0.121 | 0.293 | 18.753 |
| Experiences abdominal or pelvic pain (R10) | 0.201 | 0.154 | 0.247 | 55.018 |
| Diagnosis of other cardiac arrhythmias (I49) | 0.196 | 0.107 | 0.286 | 15.921 |
| Father diagnosed with heart disease | 0.165 | 0.141 | 0.189 | 135.459 |
| Mother diagnosed with heart disease | 0.164 | 0.137 | 0.191 | 106.746 |
| Systolic blood pressure | 0.147 | 0.135 | 0.158 | 468.358 |
| Never drinks alcohol | 0.147 | 0.11 | 0.185 | 45.945 |
| Poor self-rated health | 0.135 | 0.087 | 0.183 | 24.789 |
| Cholesterol ratio | 0.121 | 0.108 | 0.133 | 260.663 |
| Diagnosis of urinary tract infection or incontinence (I39) | 0.12 | 0.073 | 0.166 | 20.902 |
| Diagnosis of diabetes (E10, E11, E14) | 0.114 | 0.066 | 0.161 | 18.371 |
| Always adds salt to served food | 0.111 | 0.064 | 0.157 | 18.58 |
| Lost weight compared with 1 year ago | 0.108 | 0.078 | 0.138 | 39.155 |
| Regularly takes aspirin | 0.09 | 0.058 | 0.122 | 24.793 |
| Regularly takes blood pressure medications | 0.088 | 0.051 | 0.124 | 18.551 |
| Diagnosis of depressive episode (F32) | 0.087 | 0.048 | 0.126 | 16.382 |
| Has had wheeze or whistling in the chest in the last year | 0.084 | 0.057 | 0.111 | 29.853 |
| Waist-to-hip ratio | 0.081 | 0.065 | 0.096 | 78.875 |
| Number of medications taken regularly | 0.079 | 0.067 | 0.092 | 111.438 |
| Fractured bones in the last 5 years | 0.068 | 0.032 | 0.105 | 12.172 |
| Does not regularly take mineral supplements, fish oil or glucosamine | 0.062 | 0.039 | 0.085 | 23.07 |
| Number of operations | 0.055 | 0.044 | 0.066 | 76.533 |
| Pack years of smoking | 0.046 | 0.037 | 0.056 | 67.89 |
| Hours spent outdoors in winter | 0.041 | 0.031 | 0.051 | 48.062 |
| Total cholesterol | 0.027 | 0.014 | 0.04 | 14.268 |
| Smoked occasionally in the past | -0.083 | -0.117 | -0.05 | 19.724 |
| Does not have any long-standing illness, disability or infirmity | -0.088 | -0.115 | -0.062 | 33.854 |
| Has college or university degree | -0.098 | -0.125 | -0.072 | 41.719 |
| Does not take medications for cholesterol/blood pressure or insulin | -0.12 | -0.155 | -0.085 | 35.904 |
| Good self-rated health | -0.145 | -0.173 | -0.117 | 77.44 |
| Steady average usual walking pace | -0.162 | -0.197 | -0.127 | 62.637 |
| Never feels any pain or discomfort in their chest | -0.253 | -0.282 | -0.225 | 225.24 |
| Excellent self-rated health | -0.281 | -0.325 | -0.237 | 116.913 |
| Brisk usual walking pace | -0.282 | -0.322 | -0.242 | 142.583 |
| Currently does not smoke | -0.318 | -0.352 | -0.285 | 254.163 |
| Female sex | -0.491 | -0.522 | -0.459 | 678.433 |

***Supplementary Table 9: Summary of the Cox Proportional Hazards model for males.*** *The table displays coefficients = log(HR), hazard ratios and 95% confidence intervals, as well as -log2 (p-value), where p < 0.05 an null hypothesis states that the coefficient is equal to 0.*

| Covariate | log(HR) | CI log(HR) lower 95% | CI log(HR) upper 95% | -log2 (p-value) |
|---|---|---|---|---|
| Diagnosis of atrial fibrillation and flutter (I48) | 0.623 | 0.546 | 0.699 | 187.784 |
| Diagnosis of haematological cancer | 0.494 | 0.355 | 0.633 | 38.183 |
| Age | 0.486 | 0.468 | 0.505 | inf |
| Experiences dyspnoea (R060) | 0.476 | 0.317 | 0.635 | 27.701 |
| Experiences dizziness or giddiness (R42) | 0.381 | 0.21 | 0.553 | 16.19 |
| Experiences syncope or collapse (R55) | 0.322 | 0.206 | 0.438 | 24.096 |
| Regularly takes insulin | 0.302 | 0.199 | 0.405 | 26.683 |
| Diagnosis of cellulitis (L03) | 0.27 | 0.16 | 0.38 | 19.226 |
| Has diabetes-related eye disease | 0.255 | 0.14 | 0.37 | 16.136 |
| Diagnosis of epilepsy (G40) | 0.235 | 0.114 | 0.355 | 12.834 |
| Regularly takes corticosteroids | 0.205 | 0.086 | 0.323 | 10.501 |
| Sibling diagnosed with heart disease | 0.201 | 0.153 | 0.249 | 51.969 |
| Father diagnosed with heart disease | 0.187 | 0.156 | 0.219 | 101.638 |
| Diagnosis of other cardiac arrhythmias (I49) | 0.171 | 0.052 | 0.29 | 7.721 |
| Experiences abdominal or pelvic pain (R10) | 0.17 | 0.1 | 0.24 | 18.973 |
| Mother diagnosed with heart disease | 0.157 | 0.12 | 0.194 | 53.241 |
| Never drinks alcohol | 0.136 | 0.081 | 0.192 | 19.283 |
| Systolic blood pressure | 0.134 | 0.119 | 0.149 | 218.157 |
| Poor self-rated health | 0.131 | 0.067 | 0.195 | 13.944 |
| Diagnosis of diabetes (E10, E11, E14) | 0.127 | 0.068 | 0.186 | 15.354 |
| Cholesterol ratio | 0.114 | 0.099 | 0.13 | 151.602 |
| Lost weight compared with 1 year ago | 0.11 | 0.07 | 0.149 | 23.878 |
| Always adds salt to served food | 0.105 | 0.047 | 0.164 | 11.189 |
| Diagnosis of urinary tract infection or incontinence (I39) | 0.097 | 0.013 | 0.18 | 5.416 |
| Waist-to-hip ratio | 0.092 | 0.07 | 0.113 | 54.105 |
| Number of medications taken regularly | 0.075 | 0.057 | 0.093 | 50.821 |
| Regularly takes aspirin | 0.068 | 0.028 | 0.108 | 10.042 |
| Has had wheeze or whistling in the chest in the last year | 0.068 | 0.032 | 0.103 | 12.536 |
| Does not regularly take mineral supplements, fish oil or glucosamine | 0.054 | 0.024 | 0.084 | 11.081 |
| Number of operations | 0.051 | 0.035 | 0.067 | 29.981 |
| Fractured bones in the last 5 years | 0.049 | -0.002 | 0.101 | 4.038 |
| Total cholesterol | 0.049 | 0.032 | 0.066 | 24.684 |
| Regularly takes blood pressure medications | 0.042 | -0.007 | 0.092 | 3.401 |
| Pack years of smoking | 0.037 | 0.026 | 0.049 | 31.481 |
| Hours spent outdoors in winter | 0.036 | 0.024 | 0.047 | 27.946 |
| Diagnosis of depressive episode (F32) | 0.026 | -0.032 | 0.083 | 1.388 |
| Smoked occasionally in the past | -0.063 | -0.106 | -0.02 | 7.837 |
| Does not have any long-standing illness, disability or infirmity | -0.07 | -0.104 | -0.036 | 14.014 |
| Has college or university degree | -0.074 | -0.107 | -0.04 | 16.009 |
| Steady average usual walking pace | -0.13 | -0.178 | -0.082 | 23.27 |
| Good self-rated health | -0.131 | -0.167 | -0.094 | 39.015 |
| Does not take medications for cholesterol/blood pressure or insulin | -0.148 | -0.196 | -0.1 | 28.789 |
| Never feels any pain or discomfort in their chest | -0.214 | -0.251 | -0.176 | 94.816 |
| Excellent self-rated health | -0.239 | -0.295 | -0.183 | 54.267 |
| Brisk usual walking pace | -0.276 | -0.33 | -0.222 | 76.508 |
| Currently does not smoke | -0.281 | -0.323 | -0.239 | 127.893 |

***Supplementary Table 10: Summary of the Cox Proportional Hazards model for females.*** *The table displays coefficients = log(HR), hazard ratios and 95% confidence intervals, as well as -log2 (p-value), where p < 0.05 an null hypothesis states that the coefficient is equal to 0.*

| Covariate | log(HR) | CI log(HR) lower 95% | CI log(HR) upper 95% | -log2 (p-value) |
|---|---|---|---|---|
| Diagnosis of atrial fibrillation and flutter (I48) | 0.953 | 0.847 | 1.058 | 230.667 |
| Diagnosis of haematological cancer | 0.6 | 0.407 | 0.794 | 29.57 |
| Age | 0.528 | 0.504 | 0.551 | inf |
| Experiences dyspnoea (R060) | 0.443 | 0.281 | 0.605 | 23.455 |
| Experiences syncope or collapse (R55) | 0.402 | 0.253 | 0.55 | 23.109 |
| Regularly takes insulin | 0.388 | 0.246 | 0.529 | 23.619 |
| Diagnosis of epilepsy (G40) | 0.279 | 0.132 | 0.427 | 12.215 |
| Sibling diagnosed with heart disease | 0.248 | 0.199 | 0.297 | 73.59 |
| Has diabetes-related eye disease | 0.237 | 0.074 | 0.4 | 7.862 |
| Diagnosis of other cardiac arrhythmias (I49) | 0.23 | 0.094 | 0.365 | 10.139 |
| Experiences abdominal or pelvic pain (R10) | 0.225 | 0.163 | 0.288 | 38.962 |
| Diagnosis of cellulitis (L03) | 0.209 | 0.049 | 0.37 | 6.557 |
| Regularly takes corticosteroids | 0.206 | 0.081 | 0.332 | 9.636 |
| Experiences dizziness or giddiness (R42) | 0.197 | 0.002 | 0.392 | 4.405 |
| Mother diagnosed with heart disease | 0.171 | 0.132 | 0.211 | 55.857 |
| Poor self-rated health | 0.16 | 0.087 | 0.233 | 15.875 |
| Systolic blood pressure | 0.157 | 0.14 | 0.174 | 237.069 |
| Regularly takes blood pressure medications | 0.151 | 0.096 | 0.205 | 23.766 |
| Never drinks alcohol | 0.145 | 0.093 | 0.196 | 24.836 |
| Diagnosis of depressive episode (F32) | 0.142 | 0.089 | 0.195 | 22.445 |
| Regularly takes aspirin | 0.136 | 0.083 | 0.189 | 20.875 |
| Father diagnosed with heart disease | 0.136 | 0.099 | 0.173 | 40.832 |
| Always adds salt to served food | 0.131 | 0.056 | 0.205 | 10.727 |
| Cholesterol ratio | 0.117 | 0.096 | 0.139 | 87.283 |
| Diagnosis of urinary tract infection or incontinence (I39) | 0.114 | 0.058 | 0.171 | 13.675 |
| Diagnosis of diabetes (E10, E11, E14) | 0.112 | 0.03 | 0.194 | 7.12 |
| Has had wheeze or whistling in the chest in the last year | 0.11 | 0.068 | 0.152 | 21.565 |
| Lost weight compared with 1 year ago | 0.101 | 0.056 | 0.147 | 16.148 |
| Pack years of smoking | 0.082 | 0.065 | 0.1 | 64.0 |
| Number of medications taken regularly | 0.08 | 0.062 | 0.099 | 58.325 |
| Fractured bones in the last 5 years | 0.079 | 0.027 | 0.13 | 8.531 |
| Does not regularly take mineral supplements, fish oil or glucosamine | 0.069 | 0.033 | 0.105 | 12.57 |
| Waist-to-hip ratio | 0.063 | 0.04 | 0.087 | 23.31 |
| Hours spent outdoors in winter | 0.056 | 0.035 | 0.076 | 23.352 |
| Number of operations | 0.052 | 0.037 | 0.066 | 38.405 |
| Total cholesterol | 0.001 | -0.019 | 0.021 | 0.138 |
| Does not take medications for cholesterol/blood pressure or insulin | -0.099 | -0.15 | -0.048 | 12.802 |
| Does not have any long-standing illness, disability or infirmity | -0.121 | -0.163 | -0.079 | 25.95 |
| Smoked occasionally in the past | -0.133 | -0.187 | -0.079 | 19.345 |
| Has college or university degree | -0.147 | -0.19 | -0.104 | 35.176 |
| Good self-rated health | -0.162 | -0.207 | -0.118 | 40.277 |
| Steady average usual walking pace | -0.192 | -0.243 | -0.14 | 41.442 |
| Brisk usual walking pace | -0.279 | -0.339 | -0.218 | 62.934 |
| Never feels any pain or discomfort in their chest | -0.303 | -0.347 | -0.259 | 135.79 |
| Excellent self-rated health | -0.347 | -0.419 | -0.275 | 68.276 |
| Currently does not smoke | -0.364 | -0.421 | -0.308 | 119.541 |